\documentclass[conference]{IEEEtran}
\IEEEoverridecommandlockouts

\usepackage{cite}
\usepackage{amsmath,amssymb,amsfonts}
\usepackage{graphicx}
\usepackage{booktabs}
\usepackage{array}
\usepackage{multirow}
\usepackage{makecell}
\usepackage{url}
\usepackage{xcolor}
\usepackage{pifont}

\DeclareMathOperator{\TopK}{TopK}

\IEEEaftertitletext{\vspace{-1.3em}}

\begin{document}

\title{Anatomy-Privileged Distillation with Token Routing for MRI-Based Prediction of Perineural Invasion}

\author{
\IEEEauthorblockN{
Hyunsu Go$^{1}$, Youngung Han$^{1,2}$, Kyeonghun Kim$^{2}$, Junga Kim$^{1}$, Dohyun Kweon$^{2,3}$, Jinyong Jun$^{1}$, \\
Sungha Park$^{1,4}$, Anna Jung$^{1}$, Induk Um$^{5}$, Yului Jeong$^{1}$, Suah Park$^{1}$, Jina Jeong$^{1}$, Pa Hong$^{6}$, \\
Woo Kyoung Jeong$^{7}$, Won Jae Lee$^{6}$, Ken Ying-Kai Liao$^{8}$, Hyuk-Jae Lee$^{1}$, Nam-Joon Kim$^{1,{\dag}}$
}


\IEEEauthorblockA{\footnotesize
$^{1}$Seoul National University, Seoul, Republic of Korea \quad
$^{2}$OUTTA, Seoul, Republic of Korea
}
\IEEEauthorblockA{\footnotesize
$^{3}$Kyung Hee University, Seoul, Republic of Korea \quad
$^{4}$Seoul National University School of Medicine, Seoul, Republic of Korea 
}


\IEEEauthorblockA{\footnotesize
$^{5}$Chung-Ang University, Seoul, Republic of Korea \quad
$^{6}$Samsung Changwon Hospital, Changwon, Republic of Korea
}
\IEEEauthorblockA{\footnotesize
$^{7}$Samsung Medical Center, Seoul, Republic of Korea \quad
$^{8}$NVIDIA AI Technology Center, Taipei, Taiwan
}

\IEEEauthorblockA{
$^{\dag}$Corresponding author: \texttt{knj01@snu.ac.kr}
}
}

\maketitle

\begin{abstract}
Perineural invasion (PNI) is associated with poor postoperative outcomes in intrahepatic cholangiocarcinoma, but it is confirmed by surgical pathology. Existing preoperative imaging models often rely on radiologist-defined variables, contrast-enhanced imaging, or manual annotations. We propose an anatomy-privileged teacher--student framework for patient-level PNI prediction from T2-weighted MRI. During training, the teacher uses MRI with tumor and liver masks to learn dense token routing, and the student distills this guidance to retain and aggregate informative tokens under a fixed budget. Anatomical supervision is restricted to training, and the deployed model does not require masks at inference. In 155 patients, the proposed method achieved the highest mean AUROC of 0.750 among matched MRI-only baselines evaluated under the same protocol, with 1.43 GFLOPs and 8.02 ms per case on a Jetson Orin Nano Super Developer Kit.
\end{abstract}

\begin{IEEEkeywords}
computer-aided diagnosis, privileged information, transformers, embedded systems
\end{IEEEkeywords}

\section{Introduction}
Perineural invasion (PNI) is a histopathologic finding associated with tumor spread and adverse outcomes across solid malignancies~\cite{Liebig2009Cancer}. In intrahepatic cholangiocarcinoma (ICC), postoperative PNI is associated with worse recurrence-free and overall survival after curative resection~\cite{Shirai2008WJS,Zhang2020BMCCancer,Wei2022BJS}. Because PNI is confirmed by surgical pathology, preoperative assessment must rely on noninvasive data and is relevant to risk stratification before treatment~\cite{Liu2024IJS,Liu2025ClinRadiol,Qi2025WJSO}.

PNI in cholangiocarcinoma is related to interactions among cancer cells, nerves, and surrounding microenvironment~\cite{Shen2010JECCR,Faulkner2019CancerDiscov}. In ICC, nerve-related features in the tumor microenvironment and PNI-associated molecular characteristics are linked to adverse oncologic outcomes~\cite{Bednarsch2021Cancers,Meng2023HepatolInt}. MRI studies on preoperative ICC PNI prediction have also reported associations with tumor location and peritumoral or margin-related findings~\cite{Liu2025ClinRadiol,Zhang2026BMCMI,Zhou2026EJSO}. These observations suggest that the relevant imaging signal is not restricted to the lesion interior.


Several preoperative models for ICC PNI prediction have been reported, including clinicoradiologic models, CT radiomics, MRI-based nomograms, and MRI fusion or interpretable machine-learning models~\cite{Liu2024IJS,Liu2025ClinRadiol,Qi2025WJSO,Zhang2026BMCMI,Zhou2026EJSO}. Recent 3D MRI deep-learning studies have also explored localized, scale-adaptive, and attention-based architectures for noninvasive PNI prediction~\cite{Han2026Losa,Han2026MMA}. Together, these studies show that imaging contains predictive information for PNI. However, many established approaches rely on radiologist-defined variables, contrast-enhanced imaging, or explicit lesion delineation~\cite{Liu2024IJS,Liu2025ClinRadiol,Qi2025WJSO,Zhang2026BMCMI}. This creates a gap between information that can improve model development and information that is practical to require at inference.

Learning using privileged information (LUPI) addresses this gap by allowing training-time guidance that is not required at inference~\cite{Vapnik2009NeuralNetworks,LopezPaz2016ICLR}. Knowledge distillation transfers information from a teacher with richer inputs to a student with a simpler inference path~\cite{Hinton2015Distilling}. In medical imaging, this teacher--student strategy has been used to transfer information from contrast-enhanced inputs to models operating on non-contrast images~\cite{Zhao2025MedIA}. More recent medical privileged-distillation studies have used training-time information that is costly or unavailable at inference to improve routine-input models in histopathology and lesion classification~\cite{Farndale2025MedIA,Han2026MedIA}.

For patient-level PNI prediction from T2-weighted MRI, tumor and liver masks provide privileged spatial cues during training, but requiring such annotations at deployment would limit practicality. Token sparsification methods reduce transformer cost by retaining informative tokens~\cite{Rao2021NeurIPS,Liang2022ICLR}. However, standard token sparsification does not address how anatomical information available only during training should guide token retention. We therefore combine privileged distillation with token sparsification by training a mask-guided teacher to learn dense anatomy-aware routing over a 3D token grid and distilling this routing to an MRI-only student. The student retains the top-$M$ tokens and reuses the same routing scores for weighted aggregation, with computation bounded by the retained-token budget.

\begin{figure*}[t]
    \centering
    \includegraphics[width=0.84\textwidth]{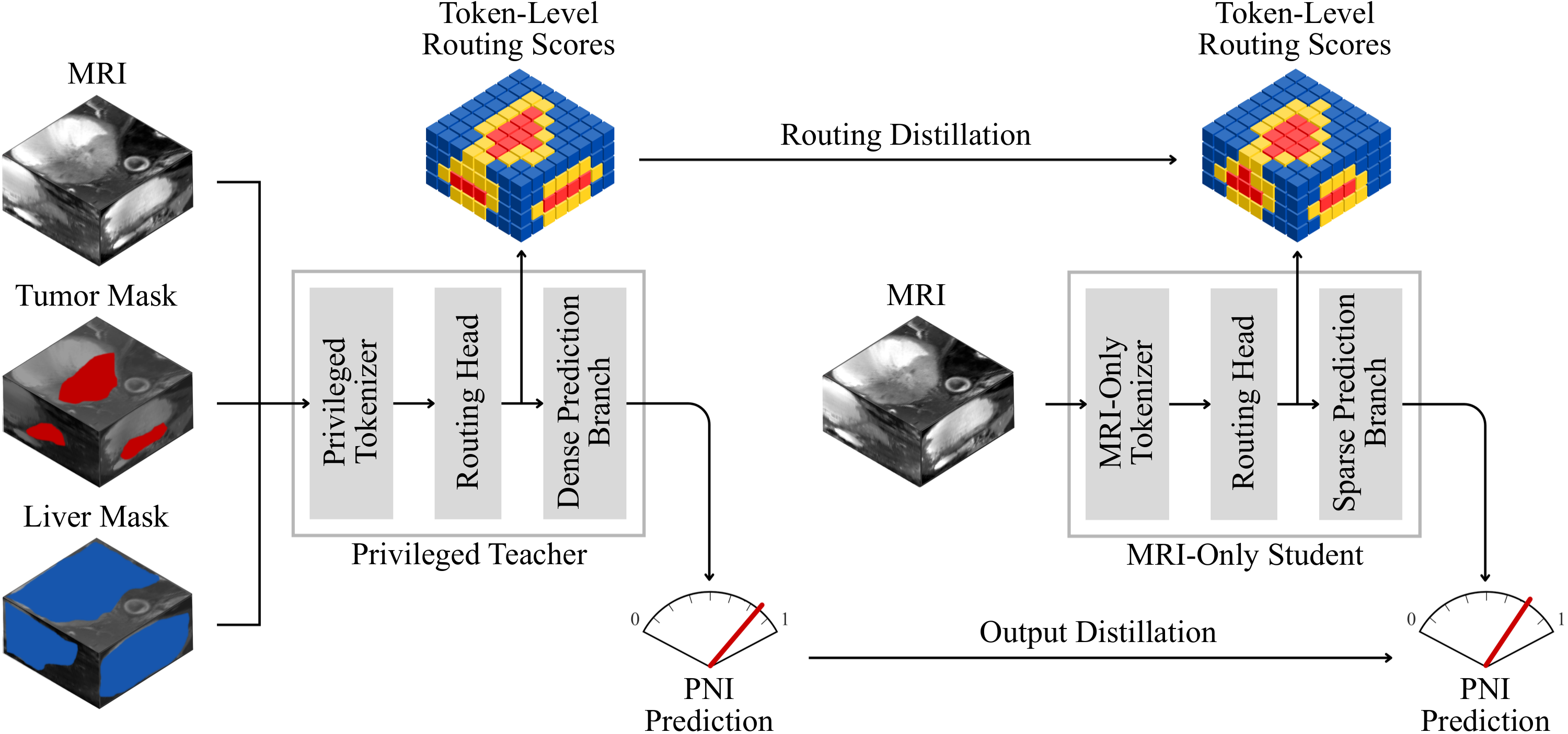}
    \caption{Overview of the proposed framework. During training, the teacher receives MRI together with tumor and liver masks, produces a dense routing distribution over the token grid, and predicts the patient-level label. The student receives MRI only and is trained with supervised learning, routing distillation, and output distillation. At inference, the student retains the top-$M$ tokens and reuses the same routing scores for weighted pooling over the retained tokens.}
    \label{fig:framework}
\end{figure*}

The main contributions of this work are as follows:
\begin{itemize}
    \item We present an anatomy-privileged teacher--student framework that transfers anatomy-aware supervision to an MRI-only classifier without requiring masks at inference.
    \item We introduce a shared-score routing design in which the same routing scores are used for both token selection and weighted aggregation of the retained tokens.
    \item We evaluate the method against matched MRI-only baselines and analyze the effects of token budget, distillation target, routing design, teacher input, and deployment cost.
\end{itemize}

\section{Proposed Method}

Let $x \in \mathbb{R}^{1 \times H \times W \times D}$ denote the input MRI volume and let $y \in \{0,1\}$ denote the patient-level PNI label. During training, privileged tumor and liver masks are available:
\begin{equation}
m_{\mathrm{tum}},\, m_{\mathrm{liv}} \in \{0,1\}^{1 \times H \times W \times D}.
\end{equation}
The teacher input is
\begin{equation}
x^{T} = [x; m_{\mathrm{tum}}; m_{\mathrm{liv}}],
\end{equation}
where $[\cdot;\cdot]$ denotes channel-wise concatenation, and the student input is
\begin{equation}
x^{S} = x.
\end{equation}
For branch $b \in \{T,S\}$, a tokenizer produces
\begin{equation}
E^{b} = \mathcal{T}^{b}(x^{b}) \in \mathbb{R}^{N \times C},
\end{equation}
where $N$ is the number of spatial tokens and $C$ is the token dimension. A routing head predicts one score per token:
\begin{equation}
s^{b} = \mathcal{G}^{b}(E^{b}) \in \mathbb{R}^{N}.
\end{equation}

\subsection{Anatomy-Privileged Teacher}

The teacher is used only during training and processes all $N$ tokens. A token mixer contextualizes the token set:
\begin{equation}
\widehat{E}^{T} = \mathcal{R}^{T}(E^{T}) \in \mathbb{R}^{N \times C},
\end{equation}
where $\mathcal{R}^{T}$ denotes the teacher token mixer. The routing scores are normalized into dense aggregation weights:
\begin{equation}
\alpha^{T} = \mathrm{softmax}(s^{T}) \in \mathbb{R}^{N}.
\end{equation}
The teacher representation and logit are
\begin{equation}
z^{T} = \sum_{i=1}^{N} \alpha_i^{T}\widehat{E}^{T}_{i} \in \mathbb{R}^{C},
\qquad
\ell^{T} = h^{T}(z^{T}),
\end{equation}
where $h^{T}$ is the teacher classifier head.


\begin{figure}[t]
    \centering
    \includegraphics[width=0.98\linewidth]{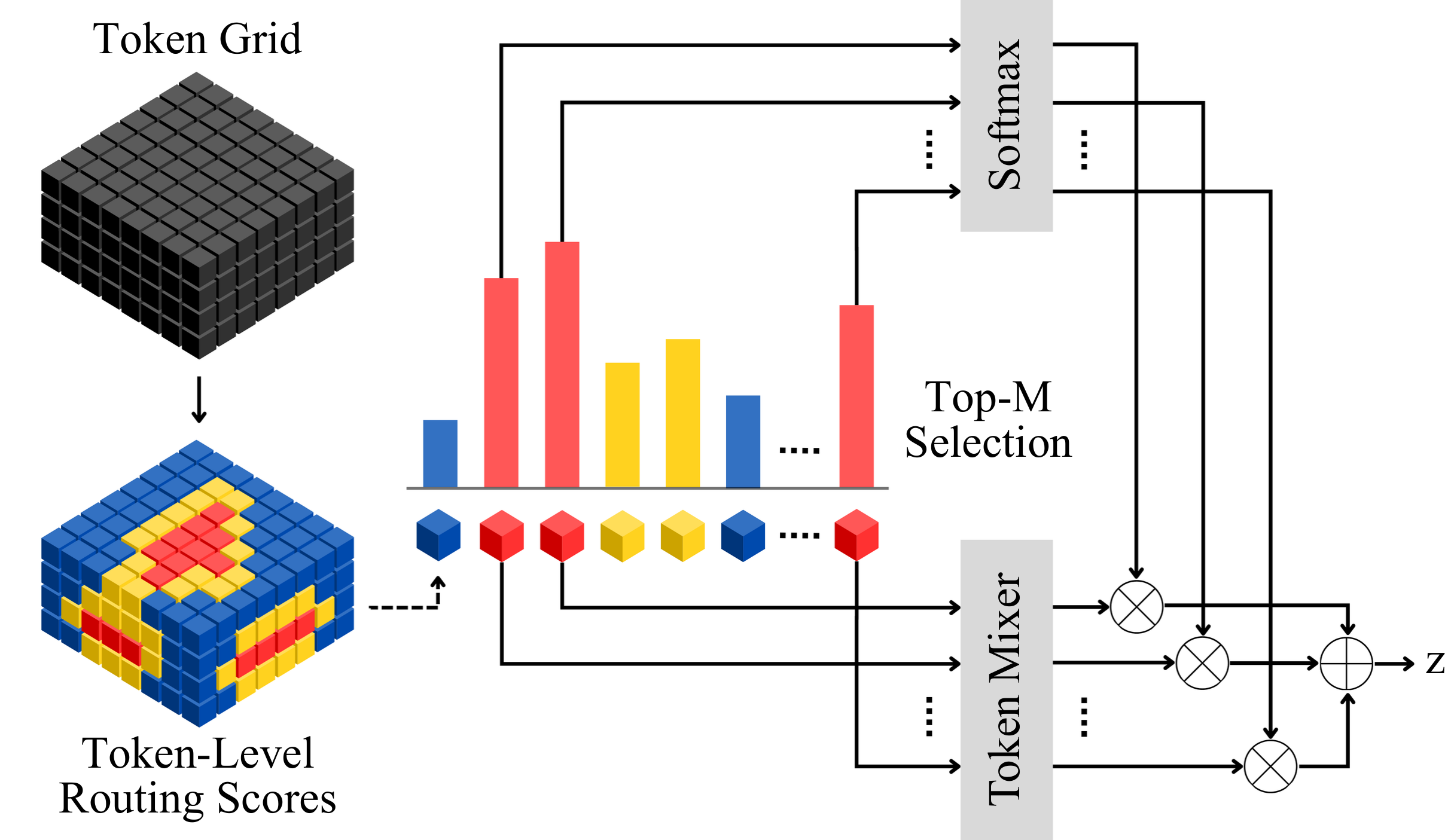}
    \caption{Shared-score routing in the student. Dense routing scores are predicted from MRI, the top-$M$ tokens are selected for token mixing, and the same retained scores are normalized for weighted pooling.}
    \label{fig:routing}
\end{figure}

\subsection{MRI-Only Student with Shared-Score Routing}

The student predicts dense routing scores from MRI only, and it processes only the top-$M$ tokens. Let
\begin{equation}
\mathcal{I}_{M} = \operatorname{Sort}\!\big(\TopK(s^{S}, M)\big),
\end{equation}
where $\TopK$ returns the $M$ highest-score token indices and $\operatorname{Sort}$ restores their original spatial order. The retained tokens and their scores are
\begin{equation}
E^{S}_{M} = E^{S}_{\mathcal{I}_{M}} \in \mathbb{R}^{M \times C},
\qquad
s^{S}_{M} = s^{S}_{\mathcal{I}_{M}} \in \mathbb{R}^{M}.
\end{equation}
Only the retained tokens are passed to the student token mixer:
\begin{equation}
\widehat{E}^{S} = \mathcal{R}^{S}(E^{S}_{M}) \in \mathbb{R}^{M \times C},
\end{equation}
where $\mathcal{R}^{S}$ denotes the student token mixer.
The same retained scores are then reused for weighted pooling:
\begin{equation}
\alpha^{S} = \mathrm{softmax}(s^{S}_{M}) \in \mathbb{R}^{M},
\end{equation}
\begin{equation}
z^{S} = \sum_{i=1}^{M} \alpha_i^{S}\widehat{E}^{S}_{i} \in \mathbb{R}^{C},
\qquad
\ell^{S} = h^{S}(z^{S}),
\end{equation}
where $h^{S}$ is the student classifier head.


\subsection{Routing and Output Distillation}

The teacher is first optimized with weighted binary cross-entropy and then frozen. For each retained-token budget $M$, the student is trained from scratch with supervised learning, routing distillation, and output distillation. 

Let
\begin{equation}
p^{T} = \sigma(\ell^{T}), \qquad p^{S} = \sigma(\ell^{S}),
\end{equation}
where $\sigma(\cdot)$ is the sigmoid function. The supervised loss for the student is
\begin{equation}
\mathcal{L}_{\mathrm{sup}}
=
- w_{+} y \log p^{S}
- (1-y)\log(1-p^{S}),
\end{equation}
where $w_{+}$ is the positive-class weight computed from the training split.

Routing distillation is applied before hard token selection so that the student receives dense supervision over all $N$ scores:
\begin{equation}
\pi^{T} = \mathrm{softmax}(s^{T}/\tau_{r}), \qquad
\pi^{S} = \mathrm{softmax}(s^{S}/\tau_{r}),
\end{equation}
\begin{equation}
\mathcal{L}_{\mathrm{route}}
=
\tau_{r}^{2}\,\mathrm{KL}\!\left(\pi^{T}\,\|\,\pi^{S}\right),
\end{equation}
where $\tau_{r}$ is the routing temperature.

For patient-level output distillation, the teacher and student logits are softened with temperature $\tau_{o}$:
\begin{equation}
q^{T} = \sigma(\ell^{T}/\tau_{o}), \qquad
q^{S} = \sigma(\ell^{S}/\tau_{o}),
\end{equation}
\begin{equation}
\mathcal{L}_{\mathrm{out}}
=
\tau_{o}^{2}
\left[
q^{T}\log\frac{q^{T}}{q^{S}}
+
(1-q^{T})\log\frac{1-q^{T}}{1-q^{S}}
\right],
\end{equation}
where $\tau_{o}$ is the output temperature.

The final student objective is
\begin{equation}
\mathcal{L}_{S}
=
\mathcal{L}_{\mathrm{sup}}
+
\lambda_{r}\mathcal{L}_{\mathrm{route}}
+
\lambda_{o}\mathcal{L}_{\mathrm{out}}.
\end{equation}

The top-$M$ operator remains discrete during both training and inference. Consequently, $\mathcal{L}_{\mathrm{sup}}$ and $\mathcal{L}_{\mathrm{out}}$ update the retained student path, while $\mathcal{L}_{\mathrm{route}}$ provides dense supervision to the routing head before pruning.

\section{Experimental Results}

\subsection{Dataset, Protocol, and Implementation}

The dataset comprised 155 patients collected over 10 years at Samsung Medical Center, including 61 PNI-positive and 94 PNI-negative cases. Only the T2-weighted MRI sequence was used for classification. Tumor and liver masks were obtained from expert annotations and were used as privileged inputs for teacher training. All methods used the same preprocessed 3D input. Volumes were resampled to a common grid, intensity-normalized over the nonzero region, and cropped to a fixed size of $96 \times 96 \times 48$.

Evaluation used a stratified five-fold cross-validation at the patient level. Reported values are cross-fold means.

The tokenizer consisted of three 3D convolution blocks with channel widths $\{32,64,128\}$, kernel size $3$, stride $2$, and GELU activation, yielding a $12 \times 12 \times 6$ token grid with $N=864$ and $C=128$. A learnable positional embedding was added before token mixing. Both teacher and student used two transformer encoder blocks with four attention heads and feed-forward hidden dimension $256$. The teacher processed all $N$ tokens, whereas the student processed only the retained $M$ tokens. Channel-wise mask dropout was applied to the privileged teacher inputs.

All models were optimized with AdamW~\cite{Loshchilov2019ICLR} for 200 epochs with batch size 4, initial learning rate $10^{-4}$, weight decay $10^{-4}$, and cosine decay after 10 warm-up epochs. The distillation temperatures were $\tau_r{=}2.0$ and $\tau_o{=}2.0$, and the loss weights were $\lambda_r{=}1.0$ and $\lambda_o{=}0.5$. Training was performed using PyTorch 2.10.0 and CUDA 12.8 on an NVIDIA RTX PRO 6000 Blackwell workstation.

Deployment profiling used batch size 1 on a Jetson Orin Nano Super Developer Kit. All profiled models used MRI only at inference. Latency, GFLOPs, and peak memory were measured on preloaded input tensors under FP16 autocast. Each latency value was averaged over 200 forward passes after 50 warm-up iterations.

\subsection{Comparison with Baselines}

We compared the proposed student with matched dense volumetric and multiple instance learning (MIL) baselines under the same input setting, fold assignment, and optimization protocol. The dense volumetric baselines were a 3D ResNet~\cite{He2016CVPR}, a 3D DenseNet~\cite{Huang2017CVPR}, and a 3D Swin Transformer implemented with a Swin UNETR-style encoder~\cite{Hatamizadeh2022BrainLes}. The MIL baselines were ABMIL~\cite{Ilse2018ICML}, DSMIL~\cite{Li2021CVPR}, and TransMIL~\cite{Shao2021NeurIPS}, adapted to the same token grid as the proposed model.

\begin{table}[t]
\caption{Patient-level comparison under the same input and evaluation protocol.}
\label{tab:comparison}
\centering
\footnotesize
\setlength{\tabcolsep}{4.0pt}
\begin{tabular}{lcccc}
\toprule
Method & AUROC & GFLOPs & \makecell{Peak mem.\\(MB)} & \makecell{Latency\\(ms)} \\
\midrule
\multicolumn{5}{l}{Dense 3D backbones} \\
ResNet-18~\cite{He2016CVPR} & 0.711 & 106.38 & 379.74 & 169.61 \\
DenseNet-121~\cite{Huang2017CVPR} & 0.726 & 18.26 & 104.00 & 42.89 \\
Swin Transformer~\cite{Hatamizadeh2022BrainLes} & 0.719 & 4.66 & 412.80 & 63.12 \\
\midrule
\multicolumn{5}{l}{MIL baselines} \\
ABMIL~\cite{Ilse2018ICML} & 0.704 & \textbf{1.43} & 27.91 & \textbf{3.08} \\
DSMIL~\cite{Li2021CVPR} & 0.673 & 1.55 & 28.17 & 3.41 \\
TransMIL~\cite{Shao2021NeurIPS} & 0.717 & 4.22 & 30.68 & 5.67 \\
\midrule
Ours ($M{=}64$) & \textbf{0.750} & \textbf{1.43} & \textbf{25.00} & 8.02 \\
\bottomrule
\end{tabular}
\end{table}

Table~\ref{tab:comparison} shows that the proposed model achieved the strongest discrimination among the compared classifiers while remaining much closer to the MIL baselines than to the dense backbones in computational cost. Relative to the dense baselines, the advantage is not limited to AUROC. The proposed model also shifts inference to a substantially lighter operating regime. Relative to the MIL baselines, it improves AUROC without increasing GFLOPs or peak memory, although the lightest MIL model remains faster in wall-clock latency.

\subsection{Ablation Studies}

\begin{table}[t]
\caption{Effect of token budget in the student. $M{=}N$ denotes the dense student.}
\label{tab:budget}
\centering
\footnotesize
\setlength{\tabcolsep}{6.0pt}
\begin{tabular}{lccc}
\toprule
\makecell{Budget $M$} & AUROC & AUPRC & GFLOPs \\
\midrule
$N$ (dense) & 0.719 & 0.604 & 4.2159 \\
128 & 0.733 & \textbf{0.714} & 1.5570 \\
96  & 0.743 & 0.706 & 1.4917 \\
64 (default) & \textbf{0.750} & 0.712 & 1.4306 \\
48  & 0.740 & 0.671 & 1.4017 \\
32  & 0.723 & 0.643 & 1.3737 \\
16  & 0.714 & 0.647 & \textbf{1.3469} \\
\bottomrule
\end{tabular}
\end{table}

Table~\ref{tab:budget} examines the effect of the token budget in the student. The budget study shows that token selection is not only a computation-saving mechanism. Moderate budgets outperform the dense student, indicating that routing can also act as feature selection by suppressing less relevant tokens before token mixing. Performance degrades when the budget becomes too small, suggesting a loss of useful context. We therefore use $M{=}64$ as the default operating point because it provides the best AUROC while remaining close to the best AUPRC with lower compute than larger budgets.

\begin{table}[t]
\caption{Ablation at $M{=}64$. In the distillation-target and aggregation-design blocks, the teacher input is fixed to MRI + tumor + liver masks. R and O denote routing distillation and output distillation, respectively.}
\label{tab:ablation}
\centering
\footnotesize
\setlength{\tabcolsep}{4.1pt}
\begin{tabular}{lcccc}
\toprule
Setting & KD & Shared & AUROC & AUPRC \\
\midrule
\multicolumn{5}{l}{Distillation target} \\
No KD & -- & \ding{51} & 0.712 & 0.638 \\
Output KD & O & \ding{51} & 0.724 & 0.643 \\
Routing KD & R & \ding{51} & 0.749 & 0.678 \\
Dual KD & R+O & \ding{51} & \textbf{0.750} & \textbf{0.712} \\
\midrule
\multicolumn{5}{l}{Aggregation design} \\
Separate pooling & R+O & -- & 0.722 & 0.657 \\
Shared-score routing & R+O & \ding{51} & \textbf{0.750} & \textbf{0.712} \\
\midrule
\multicolumn{5}{l}{Privileged teacher input} \\
MRI only & R+O & \ding{51} & 0.726 & 0.692 \\
MRI + tumor masks & R+O & \ding{51} & 0.746 & 0.708 \\
MRI + tumor + liver masks & R+O & \ding{51} & \textbf{0.750} & \textbf{0.712} \\
\bottomrule
\end{tabular}
\end{table}

Table~\ref{tab:ablation} decomposes the contribution of the distillation target, the aggregation design, and the privileged teacher input at $M{=}64$. In the distillation block, routing distillation is the primary driver of AUROC improvement, whereas output distillation alone has only a limited effect. When combined with routing distillation, however, output distillation yields an additional gain in AUPRC. This pattern is consistent with routing distillation providing the dominant spatial supervision, while output distillation refines the patient-level decision once the routing pattern is established. In the aggregation block, shared-score routing outperforms a separate pooling head under the same dual-distillation setting. This suggests that keeping token selection and token aggregation tied to the same routing signal is more effective than learning the two roles independently. In the teacher-input block, the largest gain comes from adding the tumor mask, while the liver mask provides a smaller additional improvement. This indicates that explicit lesion localization is the main source of privileged teacher guidance, with organ-level context providing complementary information.

\subsection{Qualitative Analysis}

\begin{figure}[t]
    \centering
    \includegraphics[width=1.0\linewidth]{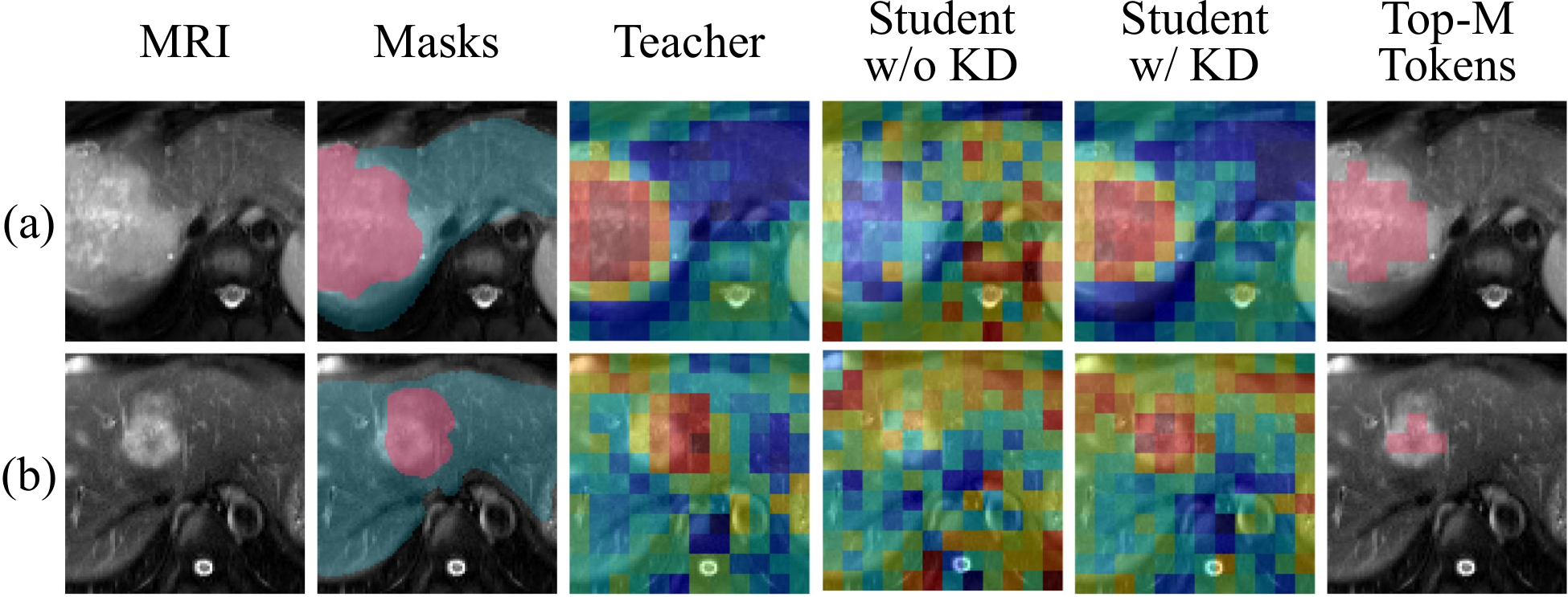}
    \caption{Qualitative routing comparison at $M{=}64$ for representative (a) PNI-positive and (b) PNI-negative patients. From left to right: MRI, masks, teacher routing, student routing without distillation, student routing with distillation, and top-$M$ retained tokens. Distillation suppresses diffuse background responses and concentrates routing on lesion-related and adjacent regions.}
    \label{fig:qualitative}
\end{figure}

Figure~\ref{fig:qualitative} provides a visual counterpart to the ablation results. Without routing distillation, the student produces more diffuse responses and allocates part of the token budget to background regions. After distillation, routing becomes more concentrated around lesion-related and adjacent regions, and the retained tokens align more closely with the teacher. This behavior is consistent with the gain observed from routing distillation in Table~\ref{tab:ablation}.

\section{Conclusion}
This work presented an anatomy-privileged teacher--student framework for patient-level PNI prediction from T2-weighted MRI. The method uses tumor and liver masks during teacher training to distill anatomy-aware routing into an MRI-only student. By linking token selection and weighted aggregation to a shared routing signal, the student enables fixed-budget inference while preserving discriminative spatial context. Under a unified evaluation protocol, the proposed model achieved the highest mean AUROC among the matched MRI-only baselines while maintaining a low deployment cost on embedded hardware. The study remains limited by the relatively small single-center cohort, moderate class imbalance, and absence of external validation. These findings suggest that anatomy-privileged routing distillation is a promising and efficient strategy for mask-free MRI-based PNI prediction.

\section*{Acknowledgment}
This work was supported by the IITP grant (IITP-2023-RS-2023-00256081) funded by MSIT, Korea, and the ANCHOR program (2026-ANCHOR-01-110) funded by the Ministry of Education and the Seoul Metropolitan Government, Republic of Korea.

\bibliographystyle{IEEEtran}
\bibliography{refs}

\end{document}